\def\year{2022}\relax
\documentclass[letterpaper]{article} 
\usepackage{aaai22}  
\usepackage{times}  
\usepackage{helvet}  

\usepackage{courier}  
\usepackage[hyphens]{url}  
\usepackage{graphicx} 
\usepackage{natbib}  
\usepackage{caption} 
\DeclareCaptionStyle{ruled}{labelfont=normalfont,labelsep=colon,strut=off} 
\usepackage{amsmath}
\usepackage{amsthm}
\usepackage{amssymb}
\usepackage{mathrsfs}

\usepackage{booktabs}
\usepackage{graphicx}
\usepackage{subfigure}
\usepackage{comment}

\usepackage{tabu}
\usepackage{algorithm}
\usepackage{algorithmic}
\usepackage{multirow}%

\usepackage{newfloat}
\usepackage{listings}
\floatstyle{ruled}
\newfloat{listing}{tb}{lst}{}
\floatname{listing}{Listing}

\newcommand{\sign}[1]{\mathrm{sgn}}
\newcommand{\argmax}[1]{\mathrm{argmax}}
\newcommand{\argmin}[1]{\mathrm{argmin}}

\title{Deep Unsupervised Hashing with Latent Semantic Components}
\author{
	Qinghong Lin\textsuperscript{\rm 1}\thanks{This work is done when Qinghong Lin is an intern at Tencent.},
	Xiaojun Chen\textsuperscript{\rm 1}\thanks{Corresponding authors},
	Qin Zhang\textsuperscript{\rm 1},
	Shaotian Cai\textsuperscript{\rm 1},\\
	Wenzhe Zhao\textsuperscript{\rm 2},
	Hongfa Wang\textsuperscript{\rm 2}
}
\affiliations{
	\textsuperscript{\rm 1}Shenzhen University, Shenzhen, China\\
	
	\textsuperscript{\rm 2}Tencent Data Platform\\
	{linqinghong@email.szu.edu.cn, \{xjchen, qinzhang\}@szu.edu.cn, cai.st@foxmail.com,\\ \{carsonzhao, hongfawang\}@tencent.com}
}

\usepackage{bibentry}

\begin{document}
\maketitle
\begin{abstract}
Deep unsupervised hashing has been appreciated in the regime of image retrieval.  
However, most prior arts failed to detect the semantic components and their relationships behind the images,
which makes them lack discriminative power.
To make up the defect, we propose a novel \textbf{Deep Semantic Components Hashing (DSCH)}, which involves a common sense that an image normally contains a bunch of semantic components with homology and co-occurrence relationships.
Based on this prior, DSCH regards the semantic components as latent variables under the Expectation-Maximization framework and designs a two-step iterative algorithm with the objective of maximum likelihood of training data.
Firstly, DSCH constructs a semantic component structure by uncovering the fine-grained semantics components of images with a Gaussian Mixture Modal~(GMM), where an image is represented as a mixture of multiple components, and the semantics co-occurrence are exploited. Besides, coarse-grained semantics components, are discovered by considering the homology relationships between fine-grained components, and the hierarchy organization is then constructed.
Secondly, DSCH makes the images close to their semantic component centers at both fine-grained and coarse-grained levels, and also makes the images share similar semantic components close to each other. 
Extensive experiments on three benchmark datasets demonstrate that the proposed hierarchical semantic components indeed facilitate the hashing model to achieve superior performance.
\end{abstract}

\begin{figure}[!h]
	\centering
	\includegraphics[width=1.0\linewidth]{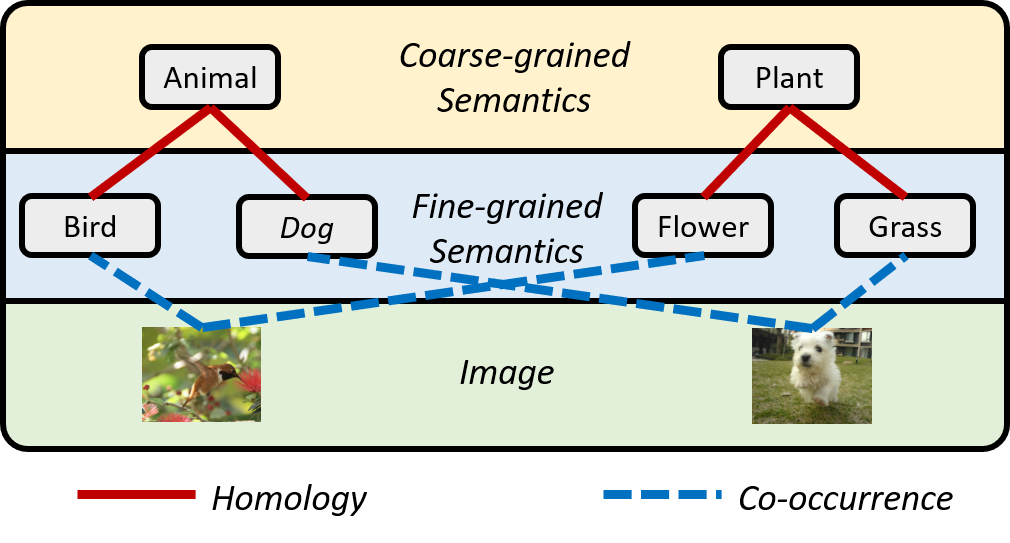}
	\caption{An illustration of semantic component structure, where a picture is normally composed of multiples interrelated semantic components. e.g, \texttt{dog} and \texttt{grass} are associated by an image, we call this relations as \textit{co-occurrence}. Besides, these semantic components are hierarchically organized from fine to coarse scale, e.g, \texttt{dog} belongs to \texttt{animal}, this relationship known as \textit{homology}.}
	\label{motivation}
\end{figure}

\section{Introduction}
With the explosive growth of social data such as images, how to conduct rapid similarity searches has become one of the basic requirements of large-scale information retrieval. Hashing~\cite{wang2015learning}~\cite{liu2016multimedia} has become a widely studied solution to this problem. The goal of hashing is to convert high-dimensional feature and similarity information into compact binary codes, which can greatly speed up computation with efficient xor operations and save storage space. 

Hashing techniques can be generally divided into supervised and unsupervised categories. Supervised hashing methods~\cite{liu2012supervised}~\cite{shen2015supervised}~\cite{yuan2020central} use label information to train hashing models, which achieve reliable performances. However, obtaining adequate labeled data is normally expensive and impractical, which hinders the application of these methods. 
In this scenario, an increasing number of researchers turn their attention to unsupervised hashing methods~\cite{liu2011hashing}~\cite{liu2014discrete}~\cite{shen2018unsupervised}~\cite{lin2021deep}.
With the lack of labels, the key to unsupervised hashing methods is the manual design of semantic similarity and how they guide the learning of hashing models.
However, existed methods~\cite{yang2018semantic}~\cite{song2018binary}~\cite{deng2019unsupervised} mainly develop the similarity information derived from the entire image, which fails to recognize the semantic composition of each image. 
Generally, a real-world picture is composed of different types of objects, which makes it contain rich semantic connotations. We call a type of object with specific semantic meaning as a \textbf{semantic component}. For instance, a picture of \texttt{walking a dog} can be comprised of the following components: \texttt{people}, \texttt{dog} and \texttt{grasses}, in which \texttt{dog} are usually correlated with \texttt{people} as human pets.
We call this phenomenon semantic \textbf{co-occurrence}.

Besides the detection of the semantic components of images and their co-occurrence relations, we notice that the \textbf{homology} relationships between them are another useful information, which is organized in a hierarchical way, such as \texttt{chihuahua} and \texttt{husky} both belong to category \texttt{dog}. This allows people to learn new concepts easily by inference its connection with learned concepts. For example, \texttt{husky} is another breed of \texttt{dog}, so it should have some similar biological qualities with \textrm{chihuahua}. The mentioned facts inspired us to introduce the homology of semantic components into the visual retrieval task. 
Several supervised hashing methods ~\cite{sun2019supervised}~\cite{wang2018supervised} proved that the semantic hierarchy can benefit hashing models. But how to model it in an unsupervised way is still an open question.

Based on the above two observations, in this paper, we propose a novel unsupervised hashing framework, called \textbf{Deep Semantic Components Hashing~(DSCH)}, which assumes that each image is composed of multiple latent semantic components and propose a two-step iterative algorithm to yield discriminative binary codes.
Firstly, DSCH constructs a semantic component structure with fully exploring the homology and co-occurrence relations of semantic knowledge, and an example is shown in Figure.~\ref{motivation}. It formulates each image as the mixture of multiple fine-grained semantic components with GMM, and clusters them to coarse-grained semantic components to construct the hierarchy. 
Secondly, DSCH makes the images close to their semantic component centers on both fine-grained and coarse-grained levels, and also makes the images share similar semantic components close to each other.
These two steps can be unified into an Expectation-Maximization framework that iteratively optimizes model parameters with the maximum likelihood of the data.
The main contributions of this paper can be summarized as follows:

\begin{itemize}
	\item We propose a novel deep unsupervised hashing framework DSCH, which treats the semantic components of images as latent variables and learns hashing models based on a two-step iterative framework.
	
	\item We propose a semantic component structure, which represents images as the mixture of multiple semantic components considering their co-occurrence and homology relationships.
	
	\item We propose a hashing learning strategy by extending the contrastive loss to adapt semantic components, where images are pulled to their semantic component centroid and close to other images with similar semantic components.
	
	\item The extensive experiments on CIFAR-10, FLICKR25K, and NUS-WIDE datasets show that our DSCH is effective and achieves superior performance.
\end{itemize}

\section{Notation and Problem Definition}
Let us introduce some notations for this paper, we use boldface uppercase letters like $\mathbf{A}$ to represent the matrix, $\mathbf{a}_{i}$ represents the $i$-th row of $\mathbf{A}$, $a_{ij}$ represents the $i$-th row and the $j$-th column element of $\mathbf{A}$ and ${\mathbf{A}}^{T}$ denotes the transpose of $\mathbf{A}$. 
$\|\cdot\|_{2}$ represents the $l_2$-norm of a vector. $\|\cdot\|_{F}$ represents the Frobenius norm of a matrix. $\sign((\cdot)$ represents the sign function, which outputs $1$ for positive numbers, or $-1$ otherwise. $\tanh(\cdot)$ represents the hyperbolic tangent function. 
$\cos(\mathbf{x},\mathbf{y})\triangleq\frac{\mathbf{x}^T\mathbf{y}}{\| \mathbf{x} \|_2 \| \mathbf{y} \|_2}$ represents the cosine similarity between vector $\mathbf{x}$ and vector $\mathbf{y}$.

Suppose we have $n$ database images $\mathbf{X}=\{\mathbf{x}_{i}\}^{n}_{i=1}$ that contains $n$ images without labels. The purpose of our method is to learn a hash function ${\mathcal{H}}: \mathbf{x}_i\rightarrow \mathbf{b}_i$ that mapping $\mathbf{X}$ into compact binary hash codes ${\mathbf{B}}=\{\mathbf{b}_{i}\}^{n}_{i=1} \in \{-1,+1\}^{n\times r}$, where  $r$ represents the length of hash codes.

\section{Related Work}
\subsection{Unsupervised Hashing}
There are many well-known traditional hashing methods been established in decades. Among them, Iterative Quantization (ITQ)~\cite{gong2012iterative} minimizes the quantization error of mapping by finding the optimal rotation of the data. 
Spectral Hashing (SH)~\cite{weiss2009spectral} construct a Laplacians graph with Euclidean distance to determine the pairwise code distances.
To speed up the construction of the graph, Anchor Graph Hashing (AGH)~\cite{liu2011hashing} proposes a sparse low-rank graph by introducing a set of anchors.
Although the aforementioned methods have made progress in this area, they are all shallow architectures that rely heavily on hand-crafted features.
To tackle this problem, amount of deep unsupervised hashing methods~\cite{deng2019unsupervised}~\cite{tu2020deep}~\cite{tu2021partial}~\cite{tu2021weighted} have been proposed, in which Deep binary descriptors (DeepBit)~\cite{lin2016learning} treats the image and its rotation as similar pairs in hash code learning.
Semantic Structure-based unsupervised Deep Hashing (SSDH)~\cite{yang2018semantic} finds the cosine distance distribution of pairs based on Gaussian estimation to construct a semantic structure. 
Twin-Bottleneck Hashing (TBH)~\cite{shen2020auto} introduces two bottlenecks that can collaboratively exchange meaningful information.
Recently,  inspired by the success in the unsupervised representation domain~\cite{he2020momentum} ~\cite{chen2020simple}, contrastive learning have been introduced to reinforce the discrimination power of binary codes~\cite{luo2020cimon}, Contrastive Information Bottleneck (CIB)~\cite{qiu2021unsupervised} modifies the contrastive loss~\cite{oord2018representation} to meet the requirement of hashing learning as:
\begin{equation}
	\mathcal{L}_0=\frac{1}{n} \sum_{i=1}^n \left(l_i^{a} + l_i^{b}\right)
	\label{l0}
\end{equation}
where	
\begin{equation}
	l_i^{a} =  -\log \frac{ e^{\cos(\mathbf{b}_i^{a}, \mathbf{b}_i^{b} )/\tau} } { e^{\cos(\mathbf{b}_i^{a}, \mathbf{b}_i^{b} )/\tau} + \sum\limits_{v, j\neq i} e^{\cos(\mathbf{b}_i^{a}, \mathbf{b}_j^{v} )/\tau}}
	\label{li}
\end{equation}
in which $\mathbf{b}_i^a=f (\mathbf{x}_i^a)$ denotes the relaxed binary codes generated from a hash encoder $f(\cdot)$ by giving a transformed view $a$ of image $\mathbf{x}_i$ as input. $v\in \{a,b\}$ denotes which view is selected and $\tau$ represents the temperature parameters. 

In the above setting, each image and its augmentation are treated as similar pairs, while all other combinations are treated as negative pairs, which will bury the similar pairs with cross samples. Therefore, in our DSCH, we expand the above objective to cover more potentially similar signals.

\subsection{Expectation-Maximum}
The Expectation-Maximum~(EM) algorithm~\cite{dempster1977maximum} has been proposed to estimate the model parameters $\Theta$ with maximum likelihood based on observed data $\mathbf{X}$ and unobserved latent variables $\mathbf{Z}=\{\mathbf{z}_j\}_{j=1}^m$:
\begin{equation}
	\sum_{i}^n \log p(\mathbf{x}_i|\Theta)=\sum_{i}^n \log \sum_{j}^m p(\mathbf{x}_i,\mathbf{z}_j| \Theta)
\end{equation}

To tackle this objective, EM contains two iterative steps:

\textbf{E step.} Expecting the posterior distribution of latent variables $\mathbf{Z}$ based on $\mathbf{X}$ and the last model parameters $\Theta^t$, which is derived as:
\begin{equation}
	Q_i(\mathbf{z}_j)=p(\mathbf{z}_j|\mathbf{x}_i, \Theta^t)
\end{equation}

\textbf{M step.} Based on the posterior distribution $Q_i(\mathbf{z}_j)$ of E step, we define the log likelihood function of $\Theta$ as
\begin{equation}
	\mathfrak{L}(\Theta) = \sum_{i}^n\sum_{j}^m Q_i(\mathbf{z}_j)\log p(\mathbf{x}_i,\mathbf{z}_j|\Theta)
	\label{M}
\end{equation}

Then, updating the model parameters $\Theta$ by maximizing the expectation of Eq.\ref{M}.
\begin{equation}
	\Theta^{t+1} = \mathop{\arg\max}_{\Theta} \mathfrak{L}(\Theta)
	\label{theta}
\end{equation}

The EM algorithm iterates the E step and the M step until convergence. By introducing the latent variables $\mathbf{Z}$, the EM algorithm has been proved to optimize the initial objective $\sum_{i}^n \log p(\mathbf{x}_i|\Theta)$~\cite{neal1998view}.

\section{Methodology}
\begin{figure}[t]
	\centering
	\includegraphics[width=0.45\textwidth]{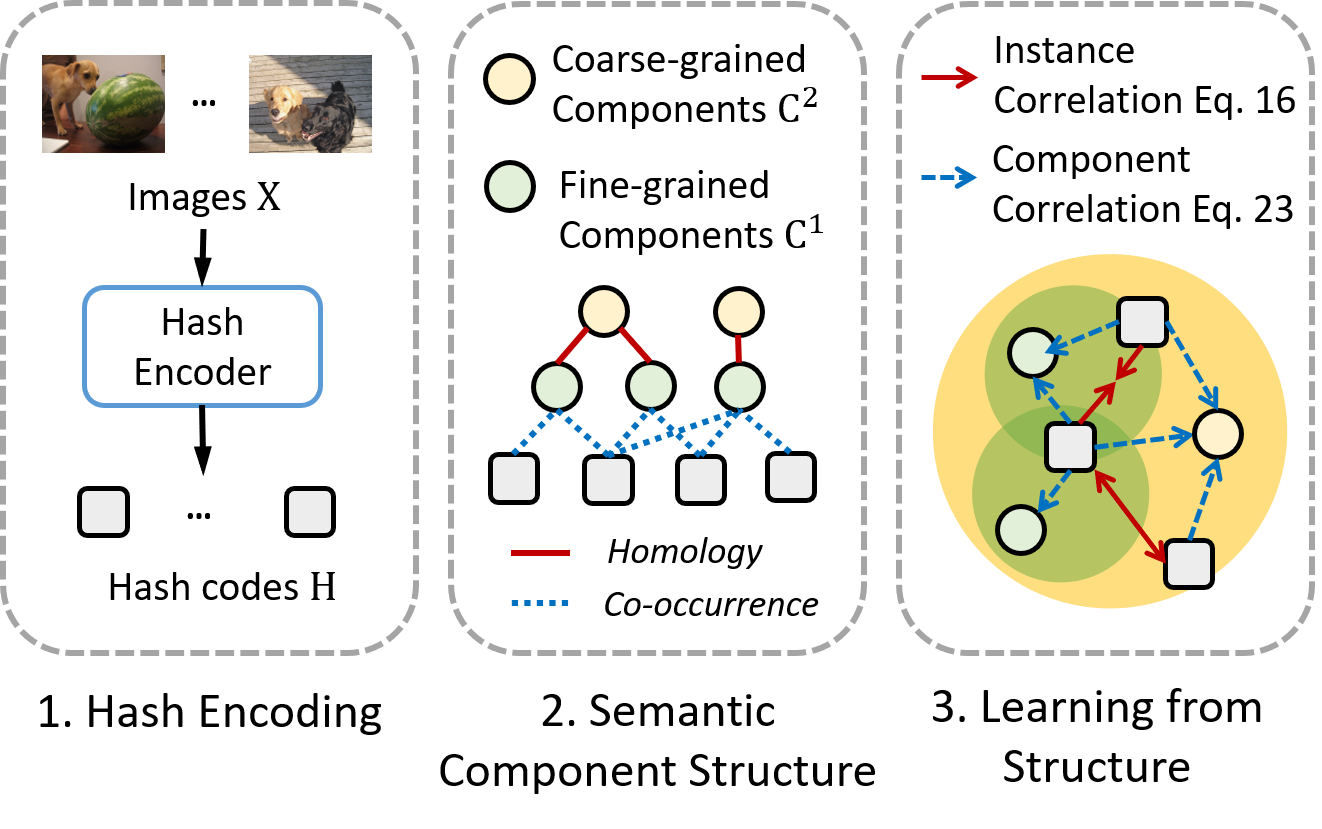}
	\caption{The pipeline of DSCH, which including three parts: (1) A hash encoder extracts the code representations for each image.  (2) Constructing a semantic component structure based on obtained code representation. (3) Optimizing the encoder by modeling two kinds of correlations.}
	\label{example}
\end{figure}

In this section, we illustrate DSCH from three folds.
Firstly, how we define the hash encoder.
Secondly, how to model the semantic components.
Lastly, how we learn binary codes from built semantic components.
An overall pipeline is shown in Fig.\ref{example}, and we will discuss each regard in detail.

\subsection{Hash Encoder}
We employ the VGG-19~\cite{simonyan2014very} as hash encoder and denote it as $f(\cdot | \Theta)$ with network parameters $\Theta$. 
To make this architecture meets the requirements of hash learning, we replace its last layers with a {fc} layer with 1000 hidden units and followed by another {fc} layer, where the node number is equal to hash codes length $r$. 
In training process, we adopt the $\tanh(\cdot)$ as the final activation function to tackle the ill-posed gradient of $\sign((\cdot)$, and get the continuous approximation of binary code $\mathbf{b}_i$ as:
\begin{equation}
	\mathbf{h}_i=\tanh \left(f(\mathbf{x}_i|\Theta) \right)\in [-1,+1]^r
	\label{hij}
\end{equation}

Once finished training, we adopt $\sign((\cdot)$ to yield the discrete binary code for Out-of-Sample extension:
\begin{equation}
	\mathbf{b}_i = \sign( \left(f(\mathbf{x}_i|\Theta)   \right) \in \{-1,+1\}^r
	\label{bij}
\end{equation}

\subsection{Semantic Component Structure}
In this subsection, we illustrate that how we build a semantic component structure with co-occurence and homology relationships.

\subsubsection{Semantic Co-occurrence}
Each image in the real world is usually composed of different types of objects, and each type of object is associated with a specific category. 
But in the unsupervised domain, since the label information is not available, it is a challenge to identify such categories meaning. 
Considering that any combination of objects may appear in an image, we assume that an image is generated from a mixture of a finite number of semantic components, where the distribution of each semantic component represents a kind of categories information.
Based on the discussion above, it is natural to adopt GMM to model the relations between data points and semantic components. 
We set the component number as a large number $m_1$ to cover any possible semantics categories in fine-grained, and we denotes these fine-grained components as $\{ C^{(1)}_j \}_{j=1}^{m_1}$. Next, we fit the parameters of components based on code representation $\mathbf{H}=\{\mathbf{h}_i\}_{i=1}^n$, therefore the distribution of a sample $\mathbf{h}_i$ can be modeled by these fine-grained components:
\begin{equation}
	p(\mathbf{h}_i)=\sum_{j=1}^{m_1} p( C^{(1)}_j)p(\mathbf{h}_i| C^{(1)}_j)=\sum_{j=1}^{m_1} \pi^{(1)}_j \mathcal{N}(\mathbf{h}_i|\mu^{(1)}_j, \Sigma^{(1)}_j)
\end{equation}
where $\pi^{(1)}_j$ is equal to $p( C^{(1)}_j)$ and satisfies $\sum_{j=1}^{m_1}\pi^{(1)}_j=1$ and $0\leq \pi^{(1)}_j \leq 1$, which denotes the prior probability of component $C^{(1)}_j$. $p(\mathbf{h}_i| C^{(1)}_j)$ is measured by a multivariate gaussian distribution $\mathcal{N}(\mathbf{h}_i|\mu^{(1)}_j, \Sigma^{(1)}_j)$ with $j$-th component parameter mean vector $\mu_j^{(1)}\in \mathbb{R}^{r}$ and covariance $\Sigma_j^{(1)}\in \mathbb{R}^{r\times r}$. These components parameters $\{\pi_j^{(1)}, \mu_j^{(1)}, \Sigma_j^{(1)}\}_{j=1}^{m_1}$ could be calculated iteratively via GMM algorithm:
\begin{equation}
	\{\pi^{(1)}_j, \mu^{(1)}_j, \Sigma^{(1)}_j\}_{j=1}^{m_1}\leftarrow \text{GMM}\left(\mathbf{H}, {m_1} \right) 
	\label{gmm}
\end{equation}

When GMM converges, we use mean vector $\mu^{(1)}_j$ to denote the representation of $j$-th fine-grained semantic component $C^{(1)}_j$ and define $\mathbf{p}_i=[p^{(1)}_{1i}, p^{(1)}_{2i}, \cdots, p^{(1)}_{m_1 i}]^T$ to represent the assignments of sample $\mathbf{h}_i$ belong to each fine-grained component, which elements $p^{(1)}_{ji}$ is estimated as:
\begin{equation}
	p^{(1)}_{ji}=p(C^{(1)}_j|\mathbf{h}_i)=\frac{\pi^{(1)}_j \mathcal{N}(\mathbf{h}_i|\mu^{(1)}_j, \Sigma^{(1)}_j) }{\sum_{k=1}^{m_1}\pi^{(1)}_k \mathcal{N}(\mathbf{h}_i|\mu^{(1)}_k, \Sigma^{(1)}_k) }
	\label{p1}
\end{equation}


\subsubsection{Semantic Homology}
Next, we explore another correlation among semantic components, name semantic homology, which means that components with similar meanings should be from the same source.
For example, \texttt{chihuahua} and \texttt{husky} are both breeds of \texttt{dogs}. Intuitively, these relationships can be organized in a hierarchical way by clustering.

We are inspired and perform $k$-means with a less cluster number $m_2<m_1$ on mean vectors $\{\mu^{(1)}_j\}_{j=1}^{m_1}$ of fine-grained components $\{ C^{(1)}_j \}_{j=1}^{m_1}$ to form coarse-grained components $\{ C^{(2)}_j \}_{k=1}^{m_2}$, where the assignment of fine-grained $C_j^{(1)}$ belongs to coarse-grained $C_k^{(2)}$ is defined as:
\begin{equation}
	p(C_k^{(2)} | C^{(1)}_j)=\left\{
	\begin{array}{l}
		1,\quad \text{if}\ C^{(1)}_j \in C_k^{(2)} \\
		0, \quad \text{otherwise}
	\end{array}
	\right.
	\label{pp}
\end{equation}

and we represent the representation of $C_k^{(2)}$ as:
\begin{equation}
	\mu^{(2)}_k = \frac{\sum_j^{m_1} p(C_k^{(2)} | C^{(1)}_j) \mu^{(1)}_j }{ \sum_j^{m_1} p(C_k^{(2)} | C^{(1)}_j) }
	\label{clij}
\end{equation}

Also, we define the prior probability of coarse-grained component $p(C^2_k)$ as $\pi^{(2)}_k=\frac{1}{m_2}$, 
since the centroid obtained with $k$-means should be treated fairly.

Lastly, based on the assignment $\mathbf{h}_i\rightarrow C^{(1)}_j$ of Eq.\ref{p1} and  $C^{(1)}_j\rightarrow C_k^{(2)}$ of Eq.\ref{pp}, we establish the assignment $p^{(2)}_{k}(i)$ of sample $\mathbf{h}_i$ to the $k$-th coarse-grained semantics $C_k^{(2)}$ by a chain rule $\mathbf{h}_i\rightarrow C^{(1)}_j\rightarrow C_k^{(2)}$, which formulated as:
\begin{equation}
	p^{(2)}_{ki}=p(C^{(2)}_k|\mathbf{h}_i)
	= \sum_{j=1}^{m_1} p(C_k^{(2)} | C^{(1)}_j) p^{(1)}_{j}(i) 
	\label{p2}
\end{equation}

\textbf{Complexity analysis.}
The complexity of constructing a semantic structure is $O(nm_1 r^3+m_1m_2 r+nm_1m_2)$ with $n$ training samples, code length $r$, $m_1$ fine-grained components and $m_2$ coarse-grained components. In the above, the first term is GMM algorithm, the second term denotes $k$-means clustering, and the third term is brought by the mapping of Eq.14. 
Notably, this complexity is linear to data samples number $O(n)$.

\subsection{Learning from Structure}
In this section, we illustrate how we learn a discriminative hash model from the built semantic component structure. We decouple this structure as multiple connections with two kinds: instance correlation and semantic correlation.
\subsubsection{Instance Correlation} 
The pairwise similarity between images is an essential cue to direct the hash model.
Generally, if two images are similar semantically, then their semantic composition should also be similar. 
For example, two pictures both contain \texttt{sky}, \texttt{dog} and \texttt{airplane} are very likely to describe a similar scene.
Thus, the similarity of images can be expressed by how similar their semantic components are. 
Based on this tuition, we develop a metric $s_{ij}$ based on sample assignments to fine-grained components $\mathbf{p}_i$ of Eq.\ref{p1} as:
\begin{equation}
	s_{ij}=\cos(\mathbf{p}_i, \mathbf{p}_j)=\frac{\mathbf{p}_i^T  \mathbf{p}_j }{\| \mathbf{p}_i \|_2 \| \mathbf{p}_j \|_2}
\end{equation}
where $s_{ij}\in [0, 1]$, denotes how similar $\mathbf{h}_i$ and $\mathbf{h}_j$ are in their distribution of assigned fine-grained semantic components. Especially, when $s_{ij}$ equals to 1, we have $\mathbf{p}_i=\mathbf{p}_j$, which means $\mathbf{h}_i$ and $\mathbf{h}_j$ are almost the same semantically. When $s_{ij}$ is equal to 0, it means $\mathbf{h}_i$ and $\mathbf{h}_j$ are totally dissimilar.

In the loss of Eq.$\ref{li}$, it normally treats two transformed views of an image as a positive pair, while ignoring the similarity information with cross samples. Thus, a straightforward idea is expanding the scope of similar pairs. We adopt the $s_{ij}$ to identify which pairs are similar and extent the contrastive loss of Eq.\ref{l0} as follows:
\begin{equation}
	\mathcal{L}_1=\sum_{i=1}^n\sum_{j=1}^n \alpha_{ij}\left( \widetilde{l}_{ij}^{aa}+ \widetilde{l}_{ij}^{ab}+\widetilde{l}_{ij}^{ba}+\widetilde{l}_{ij}^{bb} \right)
	\label{l1}
\end{equation}
where $\alpha_{ij}$ equals to $\frac{s_{ij}}{4{\sum\limits_{i} \sum\limits_{j} s_{ij}}}$ with normalization and $\widetilde{l}_{ij}^{ab}$ as:
\begin{equation}
	\widetilde{l}_{ij}^{ab}=
	-\log \frac{ e^{\cos(\mathbf{h}_i^{a}, \mathbf{h}_j^{b} )/\tau} } { \sum\limits_{v_1} \sum\limits_{v_2} \sum\limits_{i'}\sum\limits_{j'} \left( e^{\cos(\mathbf{h}_{i'}^{v_1}, \mathbf{h}_{j'}^{v_2} )/\tau} \right) }
	\label{li_new}
\end{equation}
in which $v_1\in \{a,b\}$ and $v_2\in \{a,b\}$. In the objective of Eq.$\ref{l1}$, these data pair with a higher similar semantic components $s_{ij}$ will be encouraged to be closer more with their different transformed views. Notably, our similar information is not only cross-views but also cross-samples, and the Eq.\ref{l0} can be formulated as a special case when only $s_{ii}=1$.

\subsubsection{Component Correlation}
The goal of hash encoder $f(\cdot|\Theta)$ can be formulated to find the parameters $\Theta$ with the maximum likelihood of $\sum_{i}^n \log p(\mathbf{x}_i| \Theta)$. 
By regarding these semantic components $\{ C^{(1)}_j \}_{j=1}^{m_1}$ and $\{ C^{(2)}_j \}_{j=1}^{m_2}$ as latent variables. 
We can rewrite the initial goal as:
\begin{equation}
	\sum_{i=1}^n \log p(\mathbf{x}_i|\Theta)=\sum_{i=1}^n \log \sum_{l=1}^2  \sum_{j=1}^{m_l} p(\mathbf{x}_i,C^{(l)}_j| \Theta)
	\label{em}
\end{equation}
and this objective can be solved iteratively via the EM algorithm, which includes the following two steps:

\textbf{E step.} We estimate the posterior distribution $	Q_i(C^{(l)}_j)$ of latent semantic components based on $\mathbf{X}$ and parameters $\Theta^t$, which with solutions Eq.\ref{p1} and Eq.\ref{p2}.
\begin{equation}
	\begin{aligned}
		Q_i(C^{(l)}_j)&=p(C^{(l)}_j|\mathbf{x}_i, \Theta^t)\\
		&\Rightarrow p(C^{(l)}_j|\mathbf{h}_i)=\left\{
		\begin{array}{l}
			p^{(1)}_{ji},\quad l=1 \\
			p^{(2)}_{ji}, \quad l=2
		\end{array}
		\right.
	\end{aligned}
	\label{qc}
\end{equation}

\textbf{M step.} Based on $	Q_i(C^{(l)}_j)$, we estimate the log-likelihood of $\Theta$ as:
\begin{equation}
	\mathcal{L}_2(\Theta) = \sum_{i=1}^n \sum_{l=1}^2  \sum_{j=1}^{m_l} Q_i(C^{(l)}_j) \log p(\mathbf{x}_i,C^{(l)}_j|\Theta)
	\label{theta}
\end{equation}
in which $p(\mathbf{x}_i,C^{(l)}_j|\Theta)$ equivalent to:
\begin{equation}
	\begin{aligned}
		p(\mathbf{x}_i,C^{(l)}_j|\Theta)&=p(\mathbf{x}_i|C^{(l)}_j,\Theta)p(C^{(l)}_j|\Theta)\\
		&\Rightarrow p(\mathbf{h}_i|C^{(l)}_j)p(C^{(l)}_j)
	\end{aligned}
\end{equation}
where $p(C^{(l)}_j)$ equal to $\pi^{(1)}_j$ from Eq.\ref{gmm} when $l=1$, and equal to $\pi^{(2)}_j$ when $l=2$.
The $p(\mathbf{h}_i|C^{(l)}_j)$ could be expressed as the posterior probability of sample $\mathbf{h}_i$ affiliates with $C^{(l)}_j$,
which normally relates to the cosine similarity between $\mathbf{h}_i$ and components representation $\mu^{(l)}_j$. 
Also, consider that probability $p(\mathbf{h}_i|C^{(l)}_j)$ is non-negative and should be normalized,
we formulate it as:
\begin{equation}
	p(\mathbf{h}_i|C^{(l)}_j)= \frac{e^{\cos(\mathbf{h}_i, \mu^{(l)}_j)/\tau}}{ \sum_{g=1}^{m_l} e^{\cos(\mathbf{h}_i, \mu^{(l)}_g)/\tau} }
	\label{p}
\end{equation}

Combining the Eq.~\ref{qc}, Eq.~\ref{theta} and Eq.~\ref{p} together, the solution of model parameters is defined as:
\begin{equation}
	\begin{aligned}
		\Theta^{t+1} &= \mathop{\arg\max}_{\Theta} \mathcal{L}_2(\Theta)=\mathop{\arg\min}_{\Theta} -\mathcal{L}_2(\Theta)\\
		&=\min -\sum_{l=1}^2 \sum_{i=1}^n \sum_{j=1}^{m_l} \beta^l_{ij} \log \frac{e^{\cos(\mathbf{h}_i, \mu^{(l)}_j)/\tau}}{ \sum_{g=1}^{m_l} e^{\cos(\mathbf{h}_i, \mu^{(l)}_g)/\tau} }
	\end{aligned}
	\label{l2}
\end{equation}
where $\beta_{ij}^l$ equal to $ \pi^{(l)}_j p^{(l)}_{ji}$.

The principle behind Eq.\ref{l2} is that we direct every data sample $\mathbf{h}_i$ close to their associated components center $\mu^{(l)}_j$ with its assignment weights $\beta_{ij}^k$
, with the purpose for align the instance representation with a weighted combination of components in different semantic granularities.

\subsection{Objective Function and Optimization}
We introduce data augmentation into $\mathcal{L}_2$ and integrate $\mathcal{L}_1$ into the EM framework M steps. Besides, we discretize the component center $\mu^{(l)}_j$ in Eq.\ref{l2} as $\sign( ( \mu^{(l)}_j)$ to minimize the quantization error.
Therefore, the total loss is formed as:
\begin{equation}
	\begin{aligned}
		\min_{\Theta} \mathcal{L}&= \mathcal{L}_{1} + \lambda \widetilde{\mathcal{L}_{2}}\\
		&=-\sum_{i=1}^n\sum_{j=1}^n \alpha_{ij}\left( \widetilde{l}_{ij}^{aa}+ \widetilde{l}_{ij}^{ab}+\widetilde{l}_{ij}^{ba}+\widetilde{l}_{ij}^{bb} \right)\\
		&-\lambda \sum_{v}^{a,b}\sum_{l=1}^2 \sum_{i=1}^n \sum_{j=1}^{m_l} \beta^l_{ij} \log \frac{e^{\cos(\mathbf{h}^v_i, \sign  ( ( \mu^{(l)}_j)  )/\tau}}{ \sum_{g=1}^{m_l} e^{\cos(\mathbf{h}^v_i, \sign( (\mu^{(l)}_g) )/\tau} }
	\end{aligned}
\end{equation}
where $\lambda$ is a weight coefficient of loss $\mathcal{L}_{2}$ and  $\widetilde{l}_{ij}^{ab}$ is defined in Eq.\ref{li_new}.

The algorithm of DSCH is described in Algorithm.\ref{dsch}, 
which is based on the EM algorithm, 
and its optimization contains two alternate steps:
\begin{itemize}
	\item \textbf{E step.} Sampling the semantic components $\{ C^{(1)}_j \}_{j=1}^{m_1}$ with GMM and $\{ C^{(2)}_j \}_{j=1}^{m_2}$ with $k$-means. 
	\item \textbf{M step.} Optimizing the network parameter $\Theta$ via back propagation (BP) with a mini-batch sampling. 
	\begin{equation}
		\Theta \leftarrow \Theta-\eta\nabla_\Theta(\mathcal{L}) 
		\label{bp}
	\end{equation}
	where $\eta$ is the learning rate and $\nabla_\Theta$ represents a derivative of $\Theta$.
\end{itemize}

\begin{algorithm}[!t]
	\caption{\textbf{Deep Semantic Component Hashing~(DSCH)}} 
	\begin{algorithmic}[1]
		\REQUIRE Hash model $f(\cdot | \Theta)$, image set $\mathbf{X}$, hash code length $r$, temperature factor $\tau$, semantic components number $m_1$ and $m_2$,
		weight coefficient $\lambda$, epoch $E$, learning rate $\eta$.
		\STATE Initialize parameters $\Theta^0$.
		\FOR{$t=1$ to $E$}
		\STATE $\rhd\quad $\textsc{E step.}
		\STATE Sampling $\{ C^{(1)}_j \}_{j=1}^{m_1}$ via GMM. 
		\STATE Sampling $\{ C^{(2)}_j \}_{j=1}^{m_2}$ via $k$-means. 
		\STATE $\rhd\quad$\textsc{M step.}
		\STATE Update model parameters $\Theta^{t+1}$ via Eq.\ref{bp}.
		\ENDFOR
		\STATE Obtain the $\mathbf{B}$ via Eq.$\ref{bij}$.
		\ENSURE Hash model  $f(\cdot | \Theta)$, hash codes $\mathbf{B}$
	\end{algorithmic} 
	\label{dsch}
\end{algorithm}

\section{Experiments}
In this section, we conduct experiments on various public benchmark datasets to evaluate our DSCH method.

\begin{table*}[t]
	\centering
		\begin{tabu}{l|l|lll|lll|lll}
		\toprule[1pt]
		&           & \multicolumn{3}{c|}{\textbf{CIFAR-10}} & \multicolumn{3}{c|}{\textbf{FLICKR25K}} & \multicolumn{3}{c}{\textbf{NUS-WIDE}} \\  
		Method  & Reference & 16 bits    & 32 bits    & 64 bits  & 16 bits   & 32 bits     & 64 bits  & 16 bits     & 32 bits    & 64 bits  \\ \midrule[1pt]
		LSH+VGG & STOC-02     &   0.177    &   0.192    &  0.261  &    0.596  &  0.619   &  0.650  &  0.385   & 0.455  & 0.446   \\
		SH+VGG  & NeurIPS-09    &  0.254  &  0.248     &  0.229   &    0.661  &  0.608   &  0.606  &  0.508   & 0.449  & 0.441  \\
		ITQ+VGG & PAMI-13    &  0.269  &  0.295   &  0.316   &    0.709  &  0.696   &  0.684   &  0.519   & 0.576  & 0.598   \\
		AGH+VGG & ICML-11    &    0.397   &  0.428     &  0.441  &    0.744  &  0.735   &  0.771   &  0.563   & 0.698  & 0.725  \\
		SP+VGG  & CVPR-15    &    0.280    &  0.343     &  0.365  &    0.726  &  0.705   &  0.713  &  0.581   & 0.603  & 0.673  \\
		SGH+VGG & ICML-17    &  0.286  &  0.320   &  0.347    &  0.608   &  0.657   &  0.693  & 0.463  & 0.588  & 0.638  \\ \midrule[1pt]
		GH      & NeurIPS-18    &   0.355    &  0.424     &  0.419  &    0.702  &  0.732   &  0.753  &  0.599   & 0.657  & 0.695   \\
		SSDH    & IJCAI-18   & 0.241    &  0.239   &  0.256 &    0.710  &  0.696   &  0.737 &  0.542   & 0.629  & 0.635   \\
		BGAN    & AAAI-18    &   0.535    &  0.575     &  0.587    &    0.766  &  0.770   &  0.795  &  0.719   & 0.745  & 0.761  \\
		MLS$^3$RDUH     & IJCAI-20    &   0.562   &    0.588   &   0.595   &   0.797  & 0.809  &  0.809  &   0.730 & 0.754 & 0.764  \\
		TBH     & CVPR-20    &   0.432      &  0.459     &  0.455 &    0.779  &  0.794   &  0.797  &  0.678   & 0.717  & 0.729  \\ 
		CIB     & IJCAI-21    &   0.547      &  0.583     & 0.602  & 0.773 &    0.781  &  0.798   &  0.756  &  0.777   & 0.781   \\  \midrule[1pt]
		\textbf{DSCH}    & \textbf{Proposed}  &  \textbf{0.624}   &   \textbf{0.644}  &  \textbf{0.670} &   \textbf{0.817}    &   \textbf{0.827}     &  \textbf{0.828}   &    \textbf{0.770}    &   \textbf{0.792}    & \textbf{0.801} 	\\ 
		\toprule[1pt]
	\end{tabu}
	\caption{MAP@5000 results on CIFAR10, FLICKR25K and NUS-WIDE. The best result is shown in boldface.}
	\label{sota}
	\centering
\end{table*}

\subsection{Datasets}
We evaluate our methods on three public benchmark datasets, i.e. CIFAR-10, FLICKR25K, and NUS-WIDE.

\textbf{CIFAR-10} dataset contains 60,000 images in 10 categories, where each class contains 6,000 images with size $32\times 32$. 
We randomly selected 100 images for each class as the query set, 1,000 in total. Then we used the remaining images as the retrieval set, among them, we randomly selected 1,000 images per class as the training set.

\textbf{FLICKR25K} is a dataset with multi-label, which contains 25,000 images and each image is labeled with at least one of 24 classes labels. We randomly selected 1,000 images per class as the query set and the remaining images are left for retrieval sets. In the retrieval sets, we randomly choose 10,000 images as the training set.

\textbf{NUS-WIDE} is a multi-label dataset that includes 269,648 images in 81 classes, and each image is also tagged with multiple labels more than classes. 
We selected 21 most frequent classes from the dataset and each class contains at least 5,000 related images. Among them, 2100 images are selected as a query set randomly while the remaining images were treated as a retrieval set, where 10,500 images were for training.
For multi-label datasets, if the retrieved image shares at least one label, it is regarded as being associated with the query image.

\subsection{Experiment Settings}
\textbf{Metrics.}
We evaluate model performance by three widely used metrics: Mean Average Precision (MAP) to measure the hamming ranking quality, Precision/Precision-recall curves to display model overall performance.

\textbf{Baseline Methods.}
We compared DSCH with twelve unsupervised hashing methods, including six shallow hashing models: \textbf{LSH}~\cite{andoni2006near}, \textbf{SH}~\cite{weiss2009spectral}, \textbf{ITQ}~\cite{gong2012iterative}, \textbf{AGH}~\cite{liu2011hashing}, \textbf{SP}~\cite{Xia_2015_CVPR}, \textbf{SGH}~\cite{dai2017stochastic} and six deep hashing models: \textbf{GH}~\cite{su2018greedy}, \textbf{SSDH}~\cite{yang2018semantic}, \textbf{BGAN}~\cite{song2018binary}, \textbf{MLS$^3$RDUH}~\cite{tu2020mls3rduh} and \textbf{TBH}~\cite{shen2020auto} and \textbf{CIB}~\cite{qiu2021unsupervised}. 
The parameters of the above methods referred to the setting provided by papers, and all shallow hashing methods use the {fc7} layer 4096-dimensional feature of VGG19 pre-trained on ImageNet.

\textbf{Implementation Details.}
We conducted experiments on a workstation equipped with Intel Xeon Platinum 8255C CPU and Nvidia V100 GPU.
During the training, each input image was resized to $224\times 224$. For data augmentation, we use the same strategy as CIB.
The epoch number was set to 100 with batch size 128, and the factor $\tau$ we set as 1.
We adopted the Adam optimization with learning rate $\eta$ to 5e-4. The parameters $m_1$ and $m_2$ were set to $\{1000, 1000, 2000\}$ and $\{900, 100, 1500\}$ for CIFAR, FLICKR and NUS-WIDE respectively, and $\lambda$ was fixed to 0.1 by default.

\subsection{Comparison Results and Discussions}
\begin{figure}[!t]
	\centering
	\subfigure[PR curve of CIFAR-10]{
		\includegraphics[width=0.224\textwidth, height=0.2\textwidth]{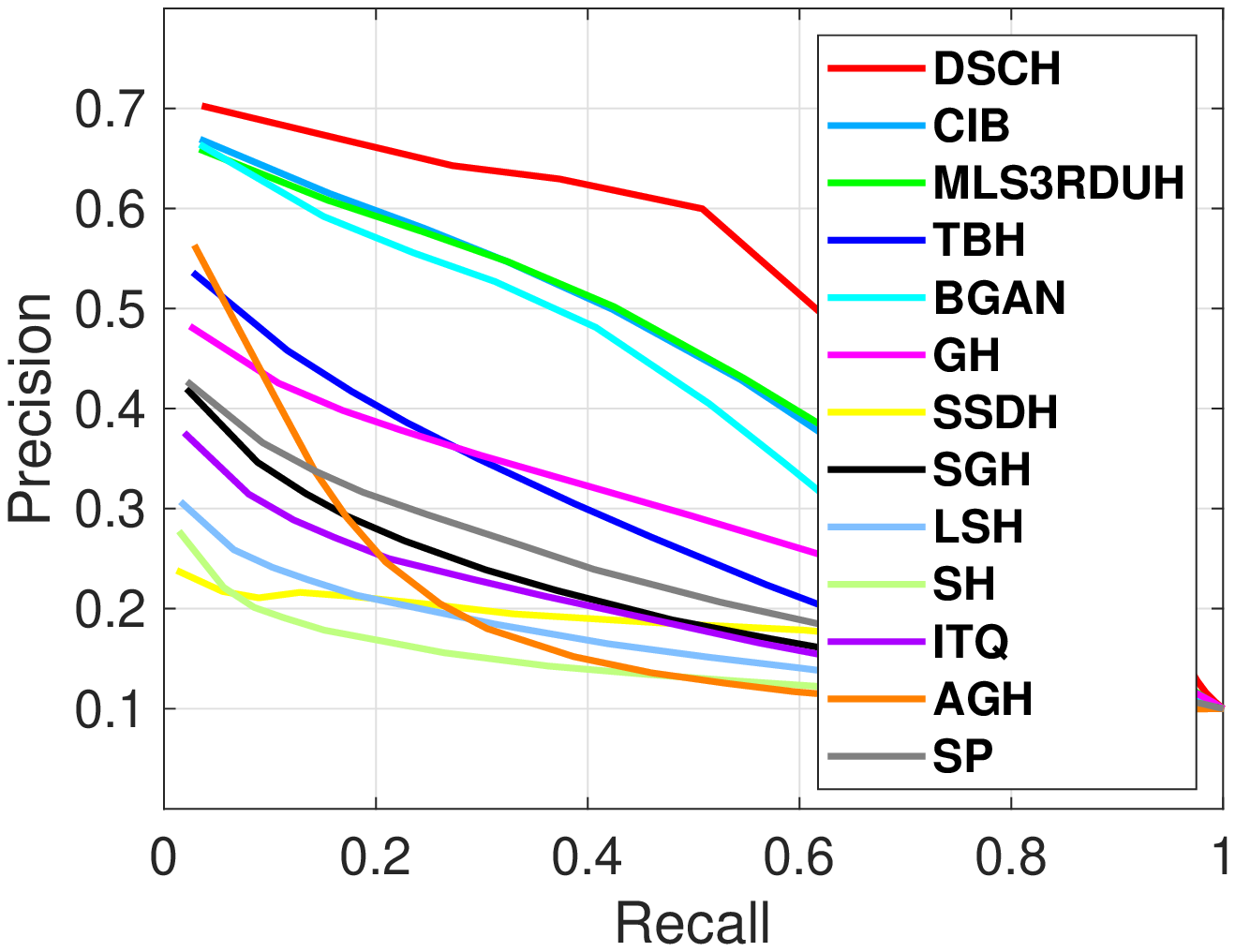}
	}
	\subfigure[Pre curve of CIFAR-10]{
		\includegraphics[width=0.224\textwidth, height=0.2\textwidth]{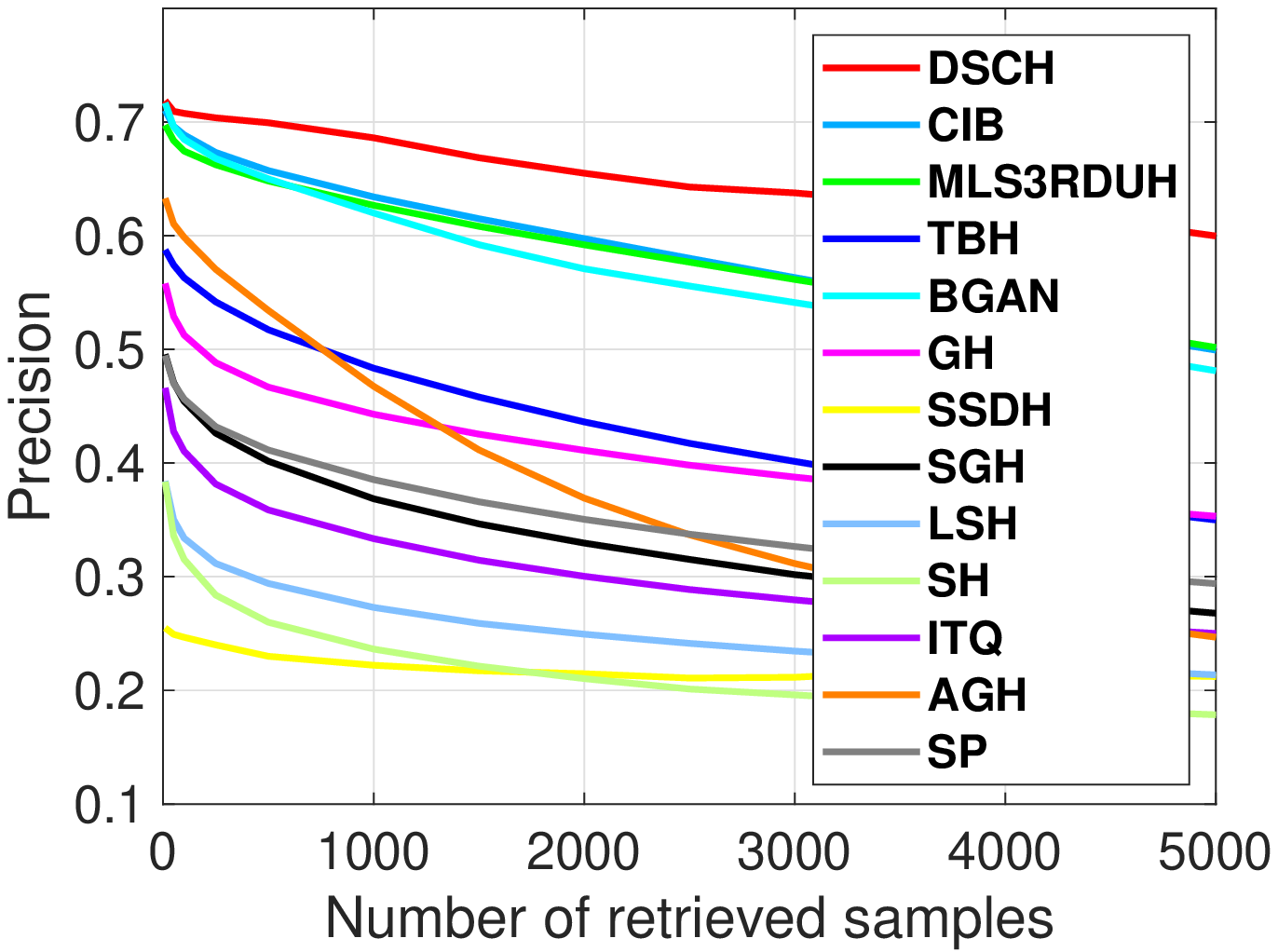}
	}
	\subfigure[PR curve of FLICKR25K]{
		\includegraphics[width=0.224\textwidth, height=0.2\textwidth]{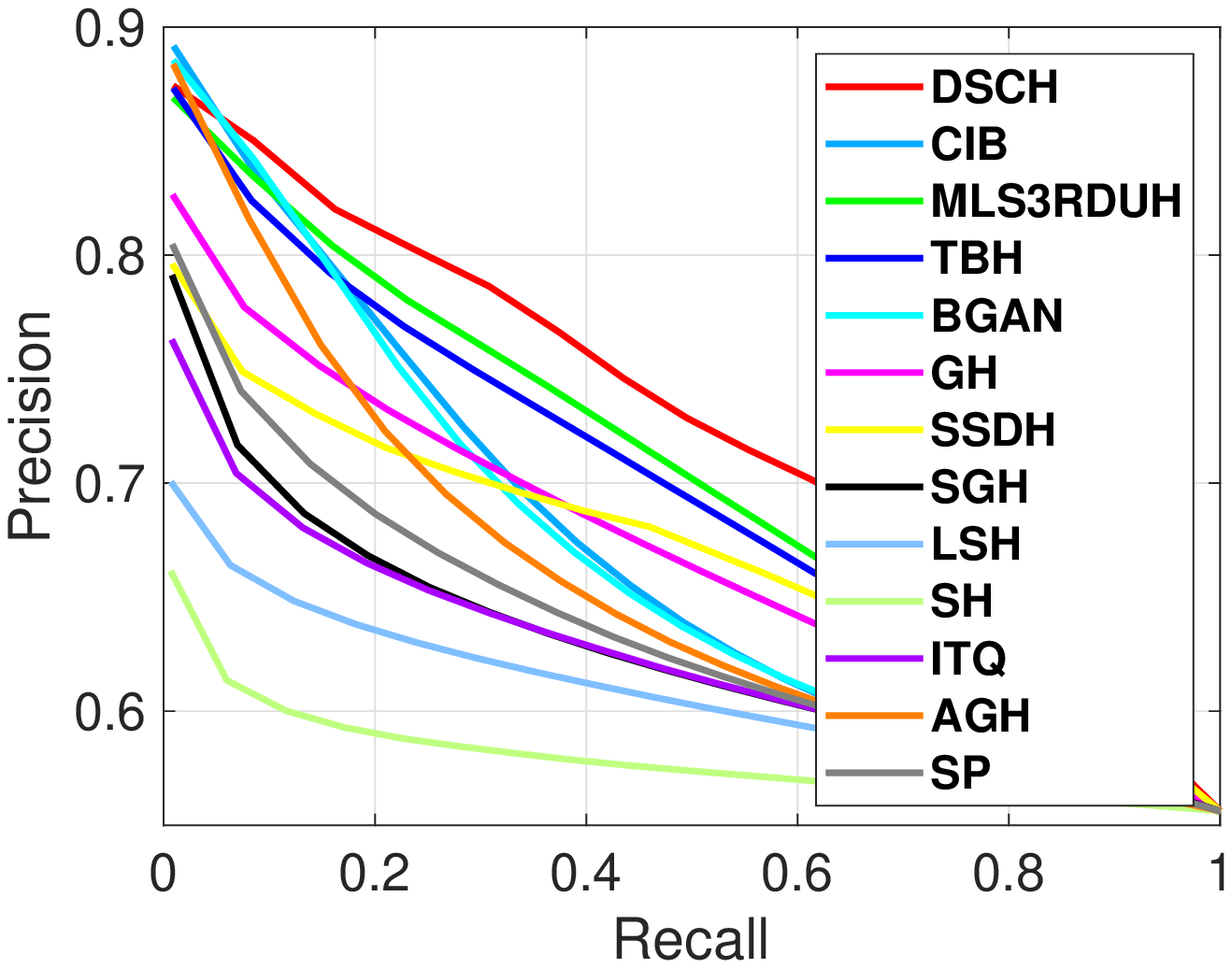}
	}
	\subfigure[Pre curve of FLICKR25K]{
		\includegraphics[width=0.224\textwidth, height=0.2\textwidth]{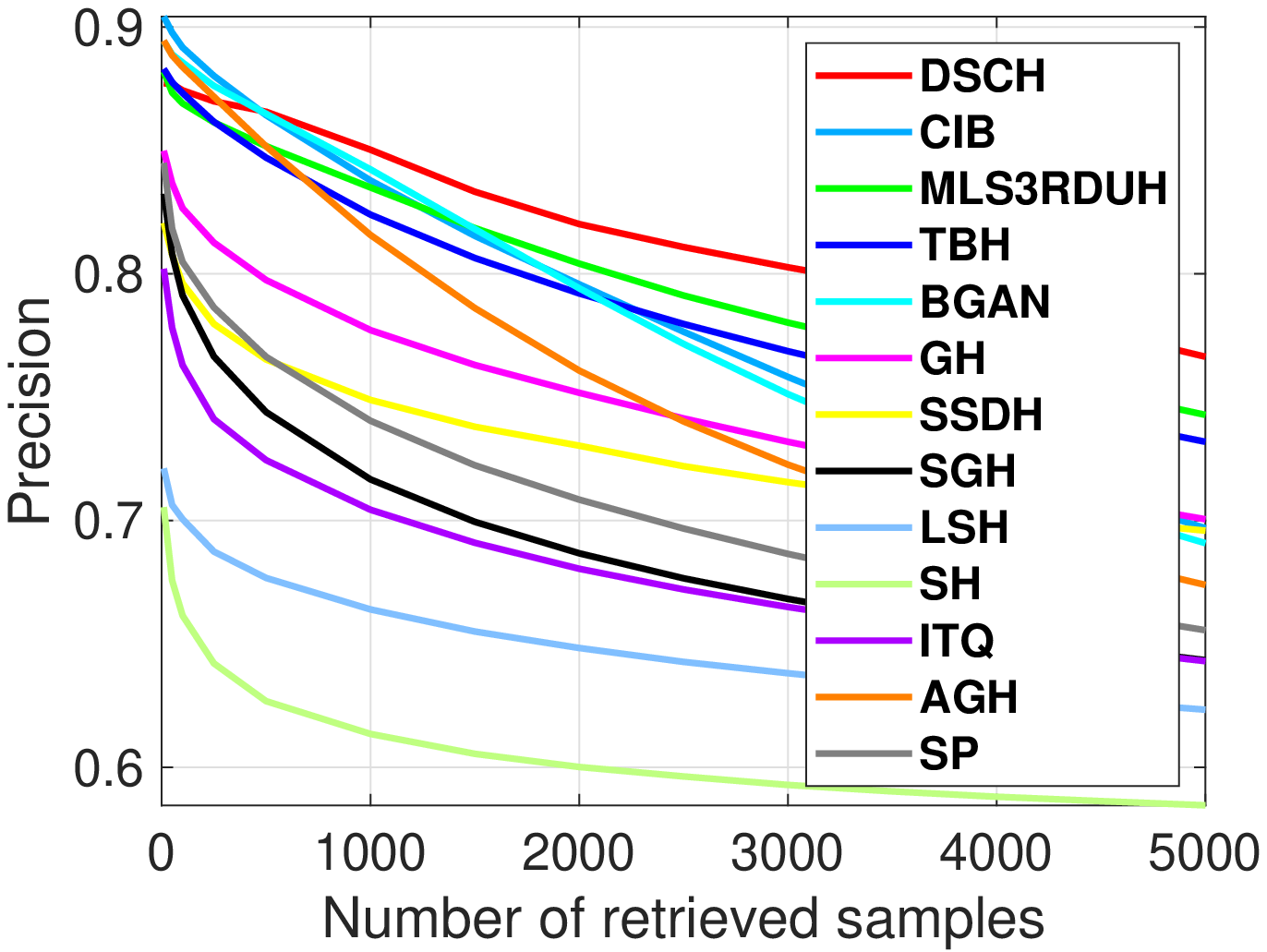}
	}
	\subfigure[PR curve of NUS-WIDE]{
		\includegraphics[width=0.224\textwidth, height=0.2\textwidth]{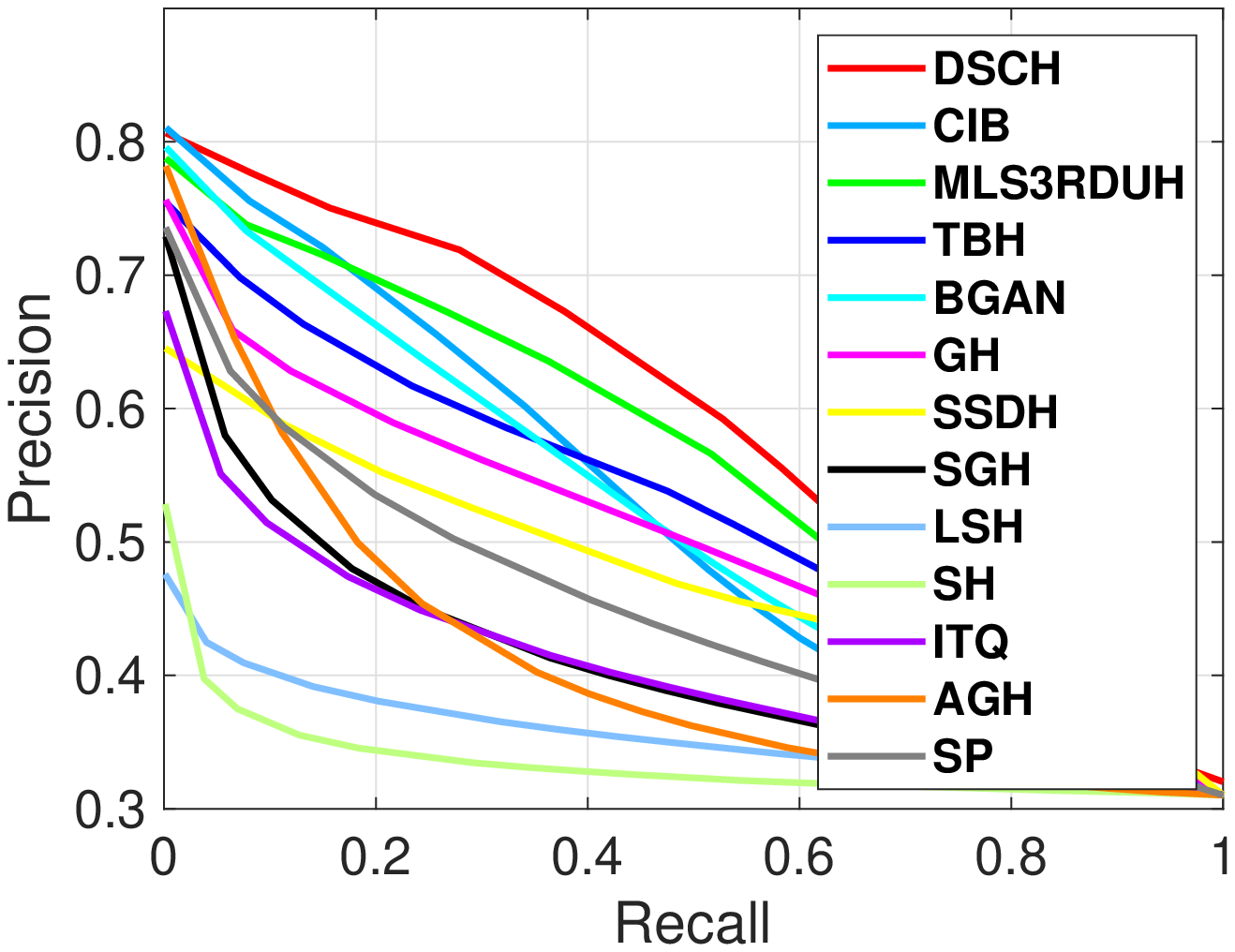}
	}
	\subfigure[Pre curve of NUS-WIDE]{
		\includegraphics[width=0.224\textwidth, height=0.2\textwidth]{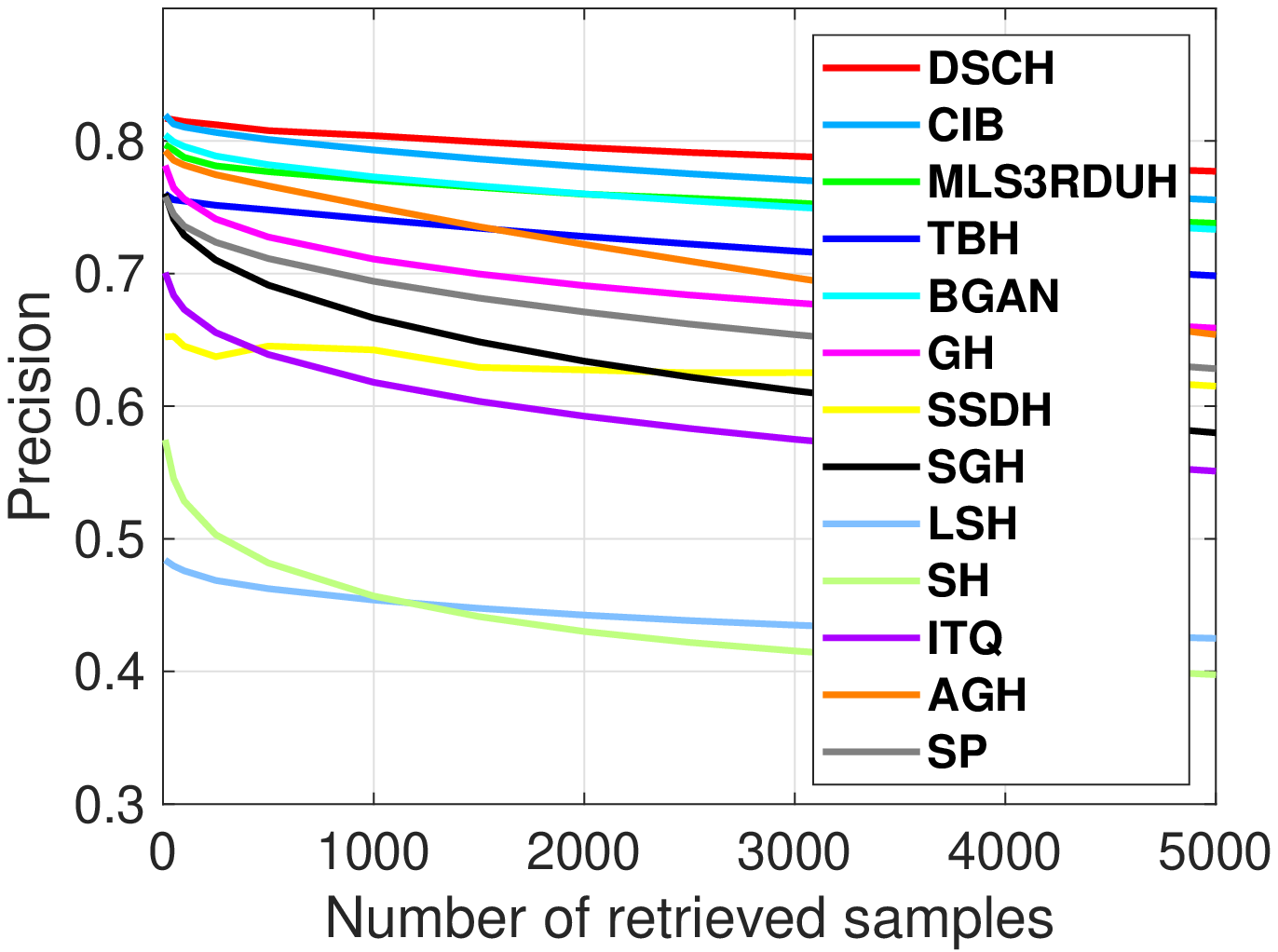}
	}	
	\caption{Precision-recall (PR) curves and Precision@5000 (Pre) curves on the CIFAR-10, FLICKR25K and NUS-WIDE datasets with 64 bits length.}
	\label{PR}
\end{figure}

The MAP@5000 results on three benchmarks are shown in Tab.~\ref{sota}, with code length varying from 16 to 64. 
It is clear that DSCH consistently achieves the best results among three datasets, with an averaged increase of 11.9\%, 5.0\%, 2.1\% on CIFAR-10, FLICKR25K, NUS-WIDE compared with the CIB. 
To sufficiently reveal the overall performance of DSCH, we report the PR curve and Precision@5000 curve of 64 bits in Fig.\ref{PR}. 
It can be found that the PR curve of DSCH covers more areas in three benchmarks and normally has a higher precision with the same returned images number. This means DSCH yields a stable superior performance.

\subsection{Ablation Study}
In this section, we conduct an ablation analysis to understand the effect of DSCH components, which consists of two major losses: 
(a) instance correlation with $\mathcal{L}_1$ by expanding the $\mathcal{L}_0$ to cover cross-sample similar pairs. 
(b) components correlation with $\widetilde{\mathcal{L}_{2}}$ to keep semantic alignment at both fine and coarse levels.
Therefore, we first design a variant with loss $\mathcal{L}_0$ as a baseline~(\texttt{Base}), and then replace it by $\mathcal{L}_1$ to evaluate the instance correlation~(\texttt{IC}). Further, equip this variant with the fine-grained~(\texttt{CC-F}) and coarse-grained components correlation~(\texttt{CC-C}) progressively.
We list the results on the FLICKR25K in Table.\ref{ab}, which are evaluated with MAP@5000 and the code length varies from 16 to 64.

\begin{table}[!h]
	\centering
	\small
		\begin{tabular}{cccc|lll}
		\toprule[1pt]
		\multicolumn{4}{c|}{\textbf{Loss Components}} & \multicolumn{3}{c}{\textbf{MAP@5000}} \\ 
		\texttt{Base}    & \texttt{IC}    & \texttt{CC-F}    & \texttt{CC-C}   & 16 bits  & 32 bits  & 64 bits \\ 	\toprule[1pt]
		$\checkmark$ &       &         &        &           0.735     & 0.744         &     0.769    \\
		$\checkmark$&   $\checkmark$    &         &        &        0.768     &          0.781 			&   0.797      \\
		$\checkmark$&   $\checkmark$    &   $\checkmark$      &        &         0.809    &          0.819			&   0.822      \\ 
		$\checkmark$&   $\checkmark$    &    $\checkmark$     &  $\checkmark$     &     \textbf{0.817}        &   \textbf{0.827}      &  \textbf{0.828}       \\ 	\toprule[1pt]
	\end{tabular}
	\centering
	\caption{Ablation study on FLICKR25K by validating the DSCH different components.}
	\label{ab}
\end{table}

As shown in the second row, introducing instance correlation is able to obtain 4.4\% (16 bits), 5.0\% (32 bits), 3.6\% (64 bits) improvements, which means that semantic components provide good clues to seek similar pairs.
Further, the performance can be gradually improved by involving the fine-grained and coarse-grained semantic correlation, resulting in 11.1\% (16 bits), 11.1\% (32 bits), 7.6\% (64 bits) gains over baseline, which indicates that the hierarchical semantic components enhance the discrimination power of model.

\subsection{Parameter Sensitivity}
\begin{figure}[t]
	\centering
	\subfigure[$(\gamma,m_2)$ in CIFAR-10]{
		\includegraphics[width=0.22\textwidth, height=0.19\textwidth]{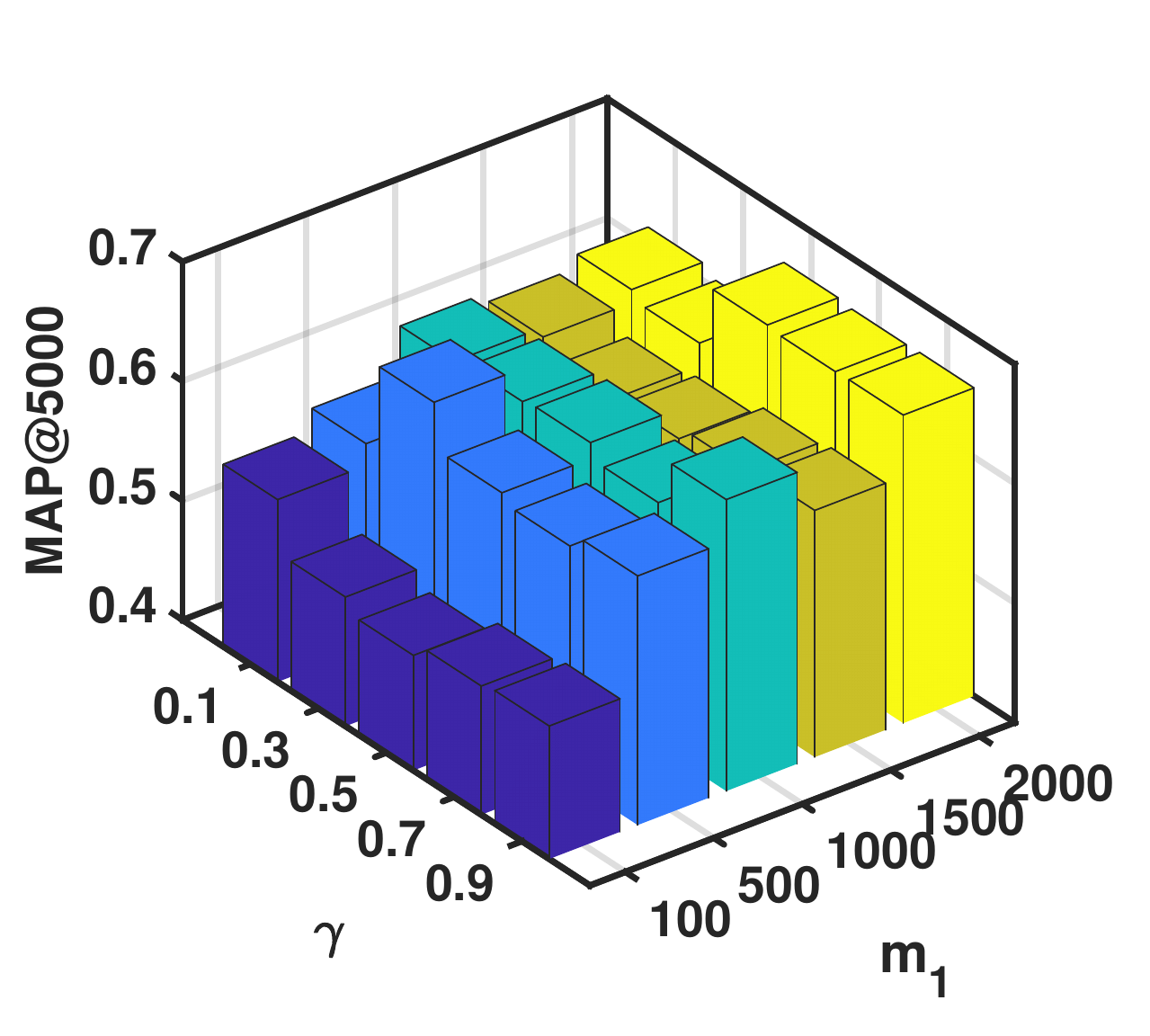}
	}
	\subfigure[$(\gamma,m_2)$ in FLICKR25K]{
		\includegraphics[width=0.22\textwidth, height=0.19\textwidth]{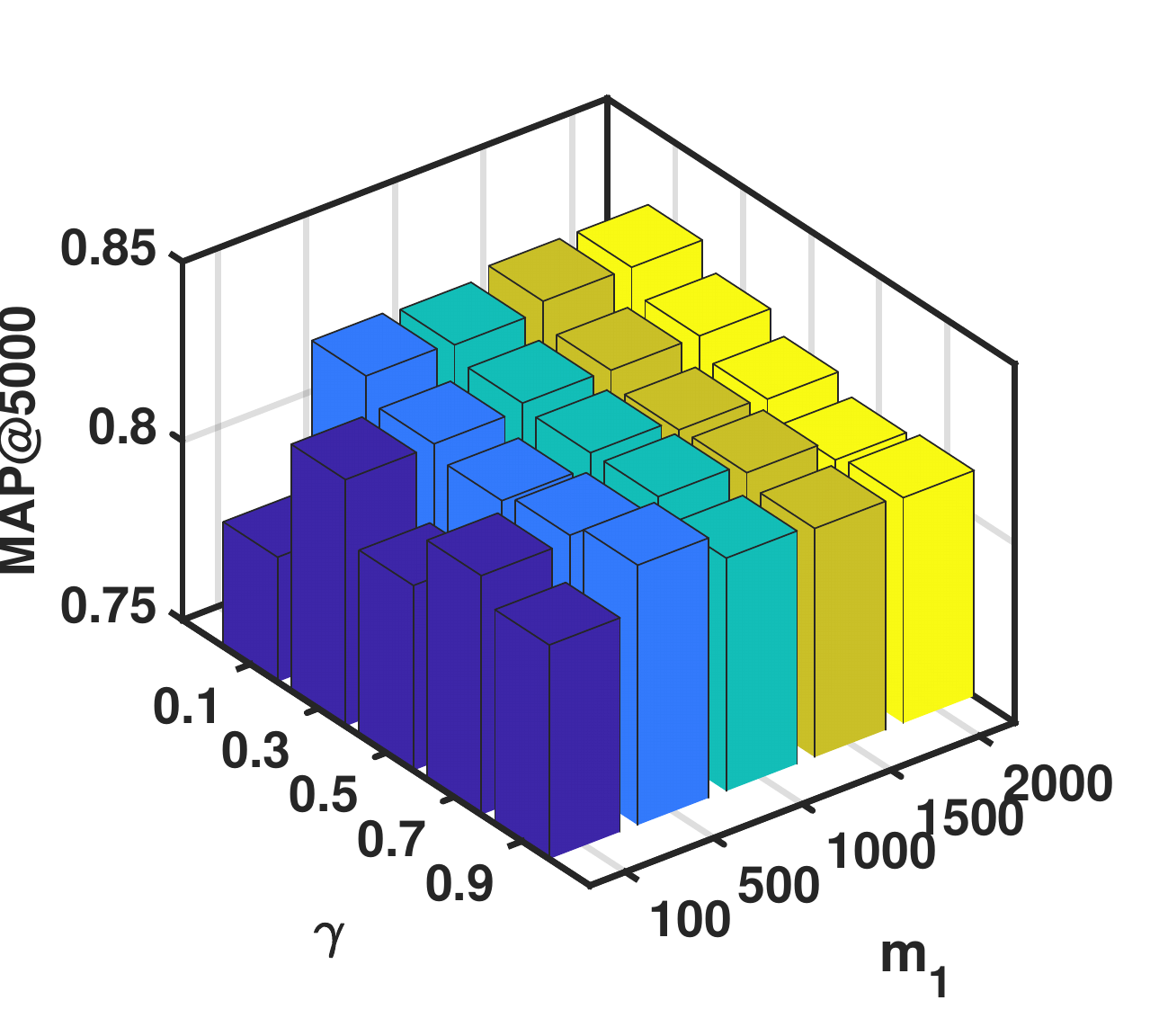}
	}
	\subfigure[$\lambda$ in CIFAR-10]{
	\includegraphics[width=0.22\textwidth, height=0.19\textwidth]{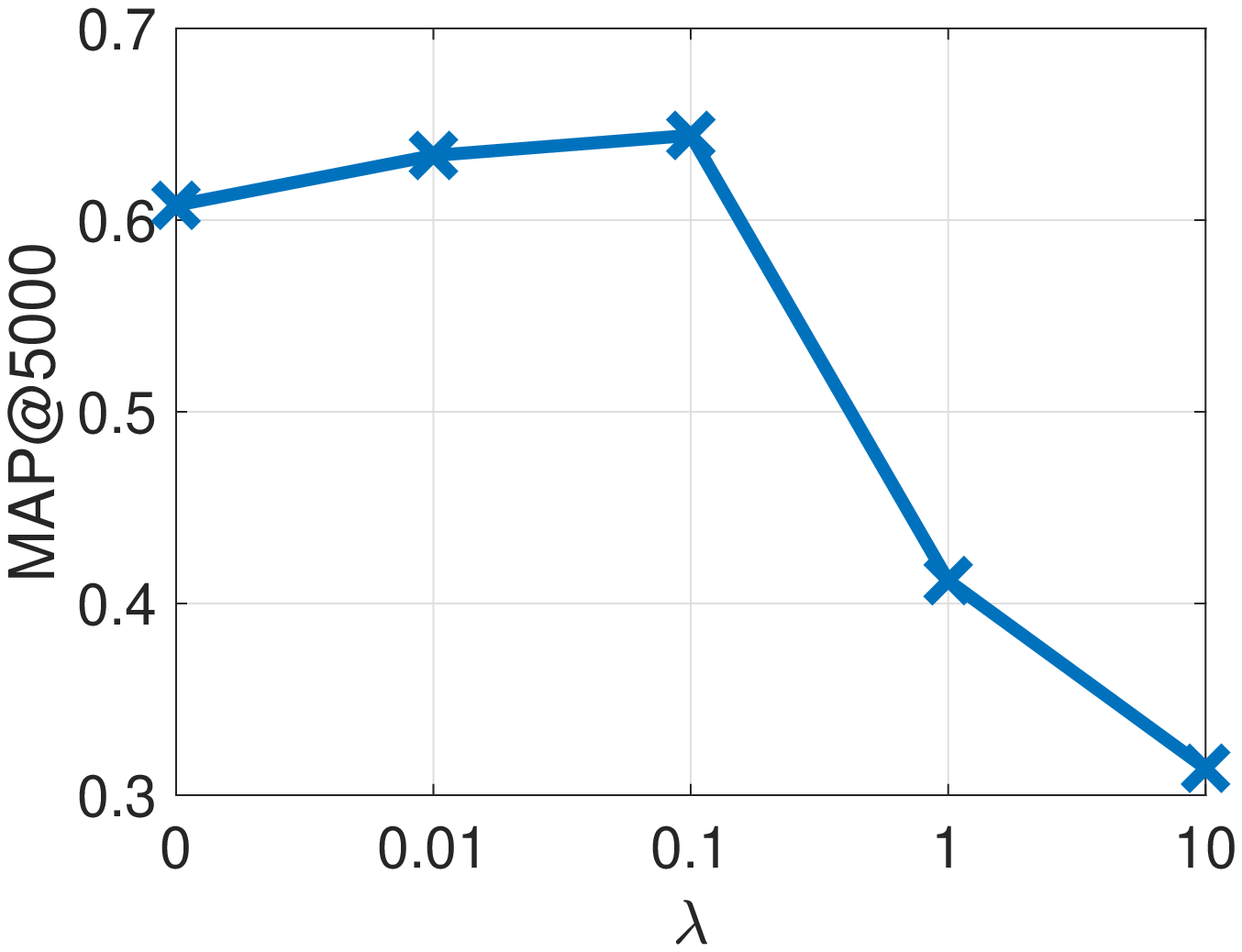}
}
\subfigure[$\lambda$ in FLICKR25K]{
	\includegraphics[width=0.22\textwidth, height=0.19\textwidth]{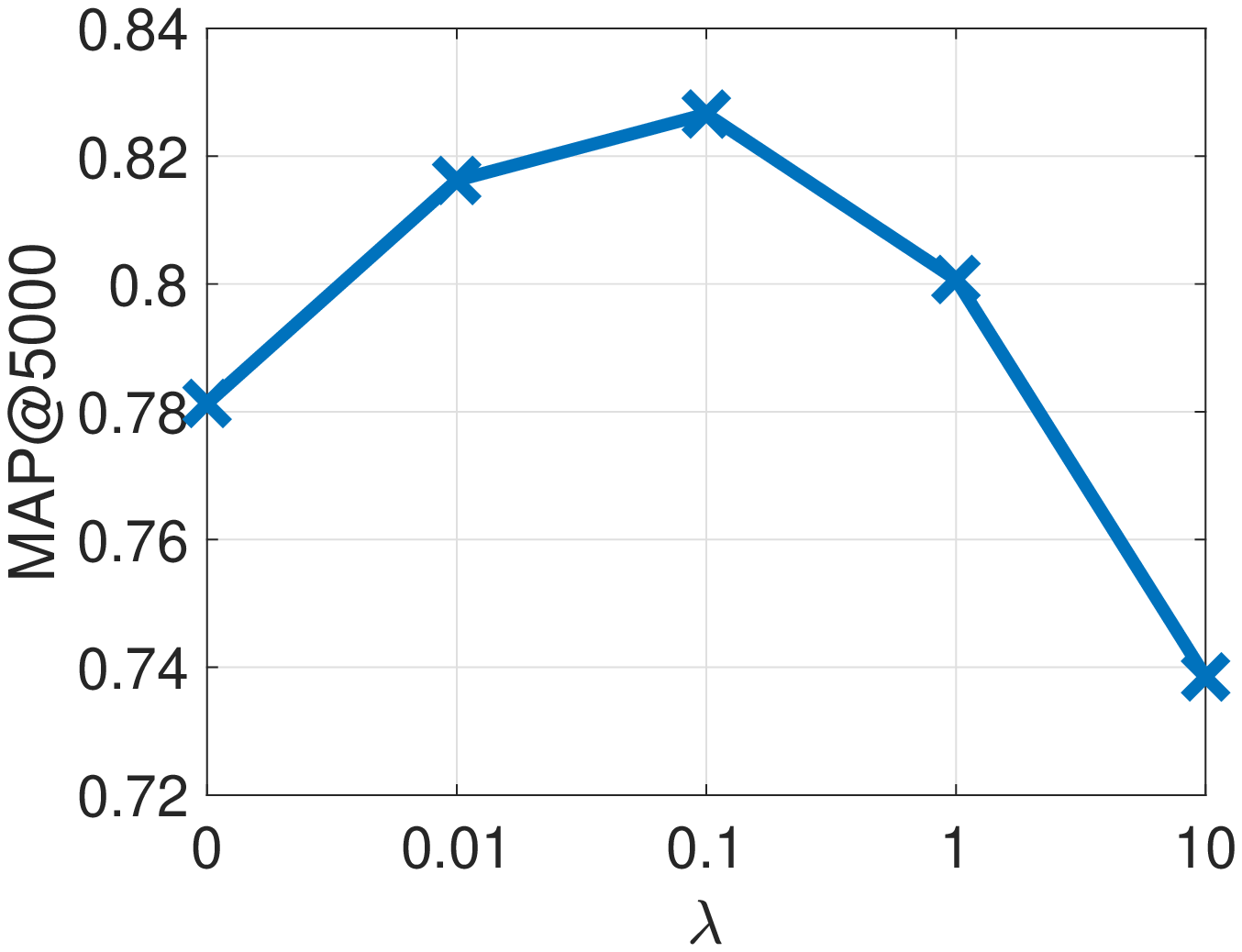}
}
	\caption{Parameter analysis on CIFAR-10 and FLICKR25K}
	\label{PA}
\end{figure}

In Fig.\ref{PA}~(a)(b), we evaluate the influence of different settings of components number $(m_1, m_2)$ on CIFAR-10 and FLICKR25K with 32 bits, where $m_1\in [100, 500, 1000, 1500, 2000]$ and $m_2=\gamma m_1, \gamma \in [0.1, 0.3, 0.5, 0.7, 0.9]$. The pattern shows that retrieval performance is jointly affected by $m_1$ and $m_2$ under different semantic granularity. 
Generally, MAP@5000 increases with $m_1$, this is due to the larger fine-grained components number bringing a finer division of semantic, which helps the model achieve better discrimination capabilities. Besides, a proper $m_2$ brings a certain performance improvement.
We noticed that in CIFAR-10, a larger $m_2$ is preferred but in FLICKR25K, a very large $m_2$ will bring relative performance degradation. 
The recommended value is $0.9m_1$ for CIFAR-10 and $0.1m_1$ for FLICKR25K.
In Fig.\ref{PA}(c)(d), we study the effect of $\lambda$, which denotes the weights of component correlation. We find that performance gain is brought when $\lambda$ smaller than $0.1$ and $\lambda=0.1 $ are a good choice in two datasets.

\subsection{Visualization Analysis}
\begin{figure}[!h]
	\setlength{\abovecaptionskip}{0.cm}
	\setlength{\belowcaptionskip}{-0.cm}
	\centering
	\subfigure[\texttt{Base}]{
		\includegraphics[width=0.14\textwidth]{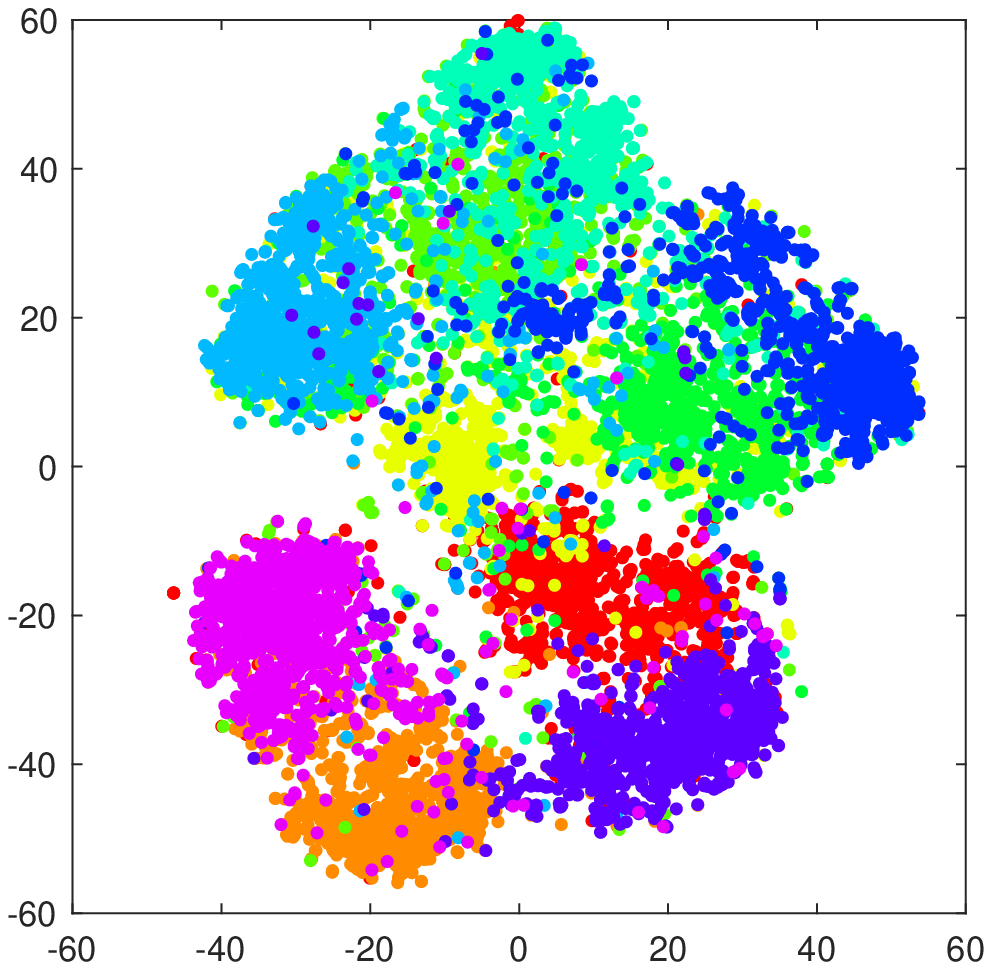}
	}
	\subfigure[\texttt{Base+IC}]{
		\includegraphics[width=0.14\textwidth]{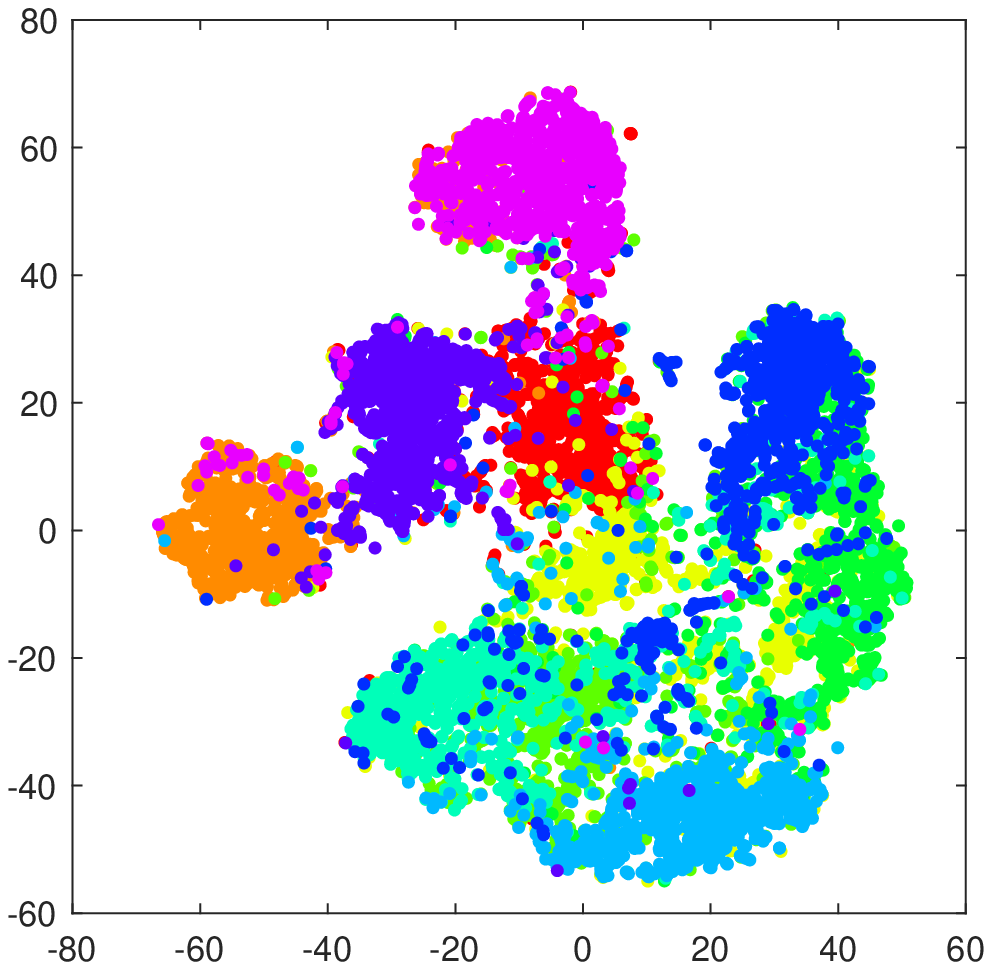}
	}
	\subfigure[\texttt{Base+IC+CC}]{
		\includegraphics[width=0.14\textwidth]{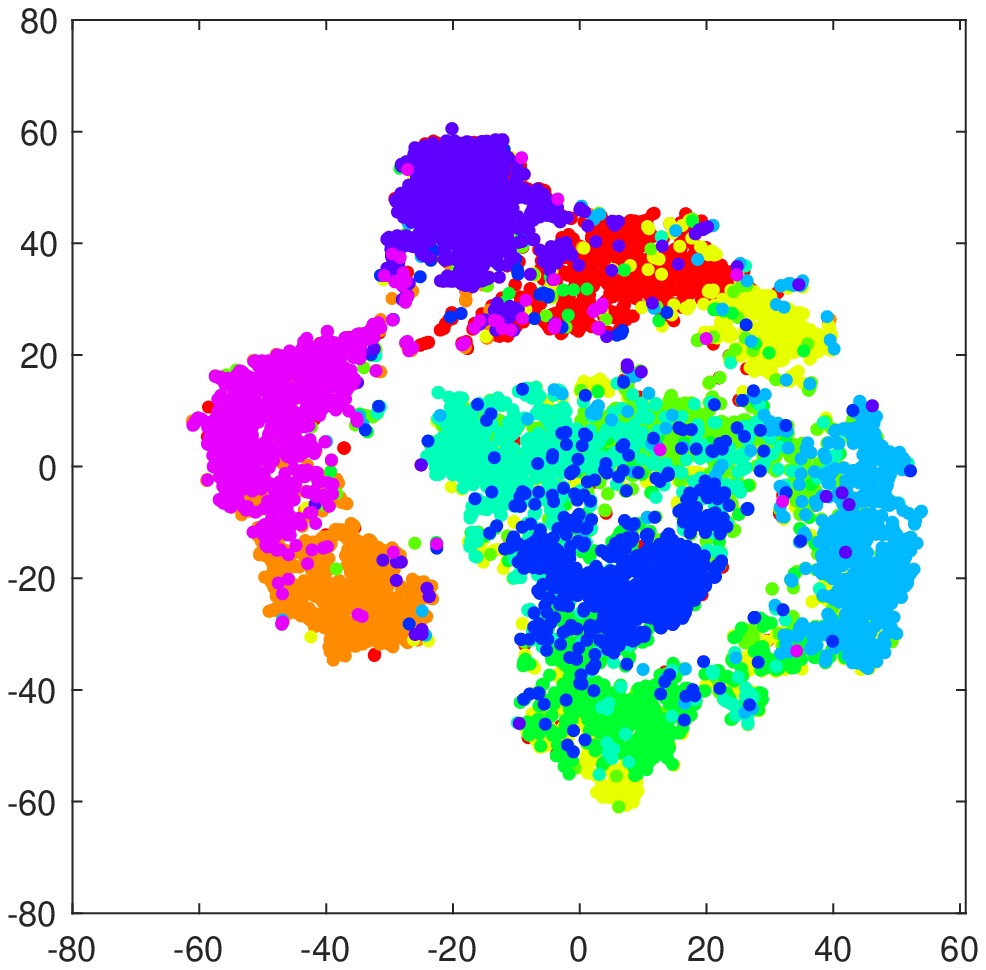}
	}
	\caption{t-SNE visualization of hash codes on CIFAR-10}
	\label{tsne}
\end{figure}

In Fig.\ref{tsne}, we display the t-SNE visualization of hash codes on CIFAR-10 with 32 bits from different variants enumerated in Tab.\ref{ab}. The color of points indicates the category samples belong to.
The figure shows that
introducing instance correlation obviously refine the manifold structure, where the images within the same class shared smaller distances to each other. 
By further equipping components correlation, data points are encouraged to be closer to their associated components, so the cluster structure is more compact.

\section{Conclusion}
We proposed a novel DSCH to yield binary hash codes by regarding every image is composed of latent semantic components and optimizing the model based on an EM framework, which includes the iterative two steps:
E-step, ~DSCH constructs a semantic structure to recognize the semantic component of images and explores their homology and co-occurrence relationships.
M-step, ~DSCH maximizes the similarities among samples with shared semantic components and pulls data points to their associated components centers at different granularities..
Extensive experiments on three datasets demonstrate the effectiveness of DSCH.

\section{Acknowledgments}
This work is jointly supported by the 2021 Tencent Rhino-Bird Research Elite Training Program; in part by Major Project of the New Generation of Artificial Intelligence (No. 2018AAA0102900); in part by NSFC under Grant no. 61773268; and in part by the Shenzhen Research Foundation for Basic Research, China, under Grant JCYJ20210324093000002.

\bibliographystyle{aaai}
\bibliography{DSCH}

\end{document}